\documentclass{article}
\usepackage{iclr2026_conference,times}

\usepackage{amsmath,amsfonts,bm}

\def\eqref#1{equation~\ref{#1}}

\def\1{\bm{1}}

\DeclareMathAlphabet{\mathsfit}{\encodingdefault}{\sfdefault}{m}{sl}
\SetMathAlphabet{\mathsfit}{bold}{\encodingdefault}{\sfdefault}{bx}{n}

\usepackage{hyperref}
\usepackage{cleveref}
\usepackage{url}
\usepackage{graphicx}
\usepackage{float}
\usepackage{placeins}
\usepackage{wrapfig}
\usepackage{booktabs,makecell}
\usepackage[table]{xcolor}

\title{Representation Distribution Matching \\ for One-Step Visual Generation}
\author{
\textbf{Lan Feng\textsuperscript{1}} \quad
\textbf{Wuyang Li\textsuperscript{1}} \quad
\textbf{Éloi Zablocki\textsuperscript{2}} \quad
\textbf{Matthieu Cord\textsuperscript{2,3}} \quad
\textbf{Alexandre Alahi\textsuperscript{1}} \\
\textsuperscript{1}EPFL, Switzerland \quad
\textsuperscript{2}Valeo.ai, France \quad
\textsuperscript{3}Sorbonne Université, France \quad
}
\newcommand{\method}{iRDM}
\newcommand{\mmmetric}{MMDr14}
\newcommand{\swmetric}{\ensuremath{\mathrm{SW}_{r^{14}}}}
\newcommand{\cvJoint}{22.76}
\newcommand{\cvMarg}{22.70}
\newcommand{\cvUntr}{19.95}

\iclrfinalcopy
\begin{document}

                                                \maketitle
\thispagestyle{plain}
\pagestyle{plain}

\begin{figure}[H]
\centering
\includegraphics[width=\textwidth]{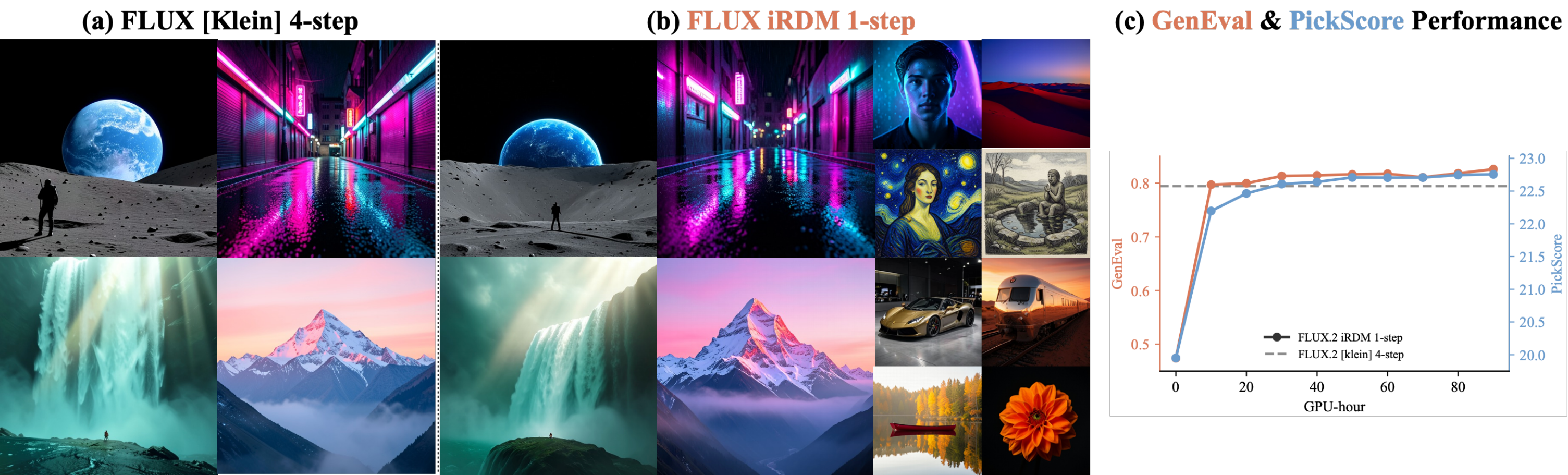}
\caption{\textbf{\method{} post-trains the four-step FLUX.2 [klein] into a one-step generator at matched quality.} (a)~Four-step FLUX.2 [klein]. (b)~One-step \method{} after post-training with the joint image-text objective.(c)~GenEval and PickScore over post-training compute, the one-step model surpassing the four-step version (grey dashed) on both metrics in about 90 H200 GPU-hours.}
\label{fig:teaser}
\end{figure}

\begin{abstract}
We elucidate the design space of Representation Distribution Matching (RDM), our name for the paradigm that trains a one-step image generator by matching generated and reference feature distributions under frozen pretrained encoders. We identify two design axes, how the distributions are compared and the representations they are compared in, and controlled studies along them yield three findings. First, the classical MMD, which could not train convincing generators a decade ago, becomes a strong and scalable objective once estimated right. Second, the generated batch is then the operative variable, with an optimum above 2048, far beyond customary batch sizes. Third, any single representation can be gamed, driven below the real score while images stay visibly fake, so we match against a balanced battery of encoders and evaluate with \swmetric{}, a Sliced-Wasserstein distance over 14 encoders that is independent of the training loss and resists gaming. Combining the preferred choices yields improved RDM (\method{}): it sets the one-step state of the art on ImageNet at \swmetric{} 1.30, corroborated by PickScore, a human-preference proxy our objective never optimizes, which prefers it over the prior best one-step generator on 71.2\% of matched samples. The same recipe post-trains the four-step FLUX.2 [klein] into a one-step generator, surpassing the four-step version on GenEval, 0.826 to 0.794, and on PickScore, \cvJoint{} to 22.58, in 90 H200 GPU-hours. Project page: https://alan-lanfeng.github.io/rdm/.
\end{abstract}

\section{Introduction}
Generative modeling is fundamentally distribution matching: we want a generator whose output distribution matches the data, and we already judge that match by the distance between their representation distributions, the basis of FID \citep{fid}. Diffusion and flow models pursue this distributional goal only implicitly, learning to reverse a noising process so that many denoising steps, simulated at inference, carry noise onto the data \citep{ddpm,scoresde,flowmatching}. A recent alternative pursues it explicitly and directly, matching the two distributions in the feature space of a frozen pretrained encoder and producing an image in a single network evaluation, with no online teacher, adversary, or trajectory to simulate. We refer to this paradigm as Representation Distribution Matching (RDM).

Several recent one-step generators \citep{drifting,fdloss} can be viewed as this paradigm, differing along just two axes. The first is the comparison: which discrepancy scores the gap between the generated and real feature laws, how it is estimated from finite samples, and what reference stands in for each side. The drifting field measures pairwise kernel forces within each batch and reads its real reference off that same batch \citep{drifting}, while the Fr\'echet-distance loss keeps only the first two moments, precomputed over the full dataset \citep{fdloss}. The second axis is the representations, by which we mean the frozen encoder feature spaces in which the two distributions are compared; here every method has settled on the same default, a few encoders under fixed weights. 

Existing methods fix these choices jointly, so it is unclear which of them is responsible for quality. We vary one axis at a time, and the resulting controlled studies overturn several assumptions implicit in current practice. 

Start with the comparison. The maximum mean discrepancy was dismissed a decade ago as too weak to train a competitive generator \citep{gmmn,dziugaite}; it was never too weak, only badly estimated. A good estimate needs a structured feature space and enough samples on each side, and the two sides differ. The reference is fixed in advance and never moves, so we use all of it: the entire 1.28M-image training set is compressed once into a frozen Nystr\"om reference \citep{nystromkme}, 4096 landmarks standing in for the attraction at a fraction of the cost (\cref{fig:toy}). The generated side moves at every step and is drawn fresh, where a larger batch sharpens the estimate but buys fewer updates; the optimum lies above 2048, an order of magnitude past common practice, with gradient caching \citep{gradcache} absorbing the memory. Finally, on conditional tasks we match the joint law of caption and image features rather than the image marginal alone, making prompt fidelity part of the objective, which post-trains four-step FLUX.2 into a one-step model at a higher GenEval.

Now the representations. Modern pretrained encoders already provide good spaces in which to measure the distance, so the question is which space, or which combination of spaces, makes a low MMD achievable only by genuinely realistic samples. A single encoder is not enough: the generator overfits whichever one it trains against, beating the real data on that encoder's own score while its samples stay visibly fake. The fix is to rely on none alone. We match across a diverse battery of encoders, and rather than weight them uniformly we keep them in balance by constrained optimization: a proportional Lagrangian controller \citep{pidlag} upweights whichever encoder is hardest to satisfy and downweights whichever the generator is beginning to overfit. The intuition is the weakest-stave rule: just as a bucket holds water only to its shortest stave, a viewer judges an image by its most pronounced artifact \citep{mad,iqapooling}, so the encoder that still objects is the one worth heeding.

Combining the two axes gives improved RDM (\method{}), a simple but effective recipe that generates in a single step at higher quality. We measure it with our new metric \swmetric{}, a relative Sliced-Wasserstein distance averaged over 14 pretrained encoders, with real data scaled to 1. As an evaluation metric the Sliced-Wasserstein distance is harder to game than the Fr\'echet distance or the MMD \citep{mind}, and since we never train against it, a gain rules out reward hacking. Post-training pMF-H FD-SIM~\citep{pmf,fdloss}, whose \swmetric{} result held the previous state of the art 2.05, \method{} reaches a new one-step state of the art at \swmetric{} 1.30, corroborated by a $71.2\%$ PickScore \citep{pickscore} win rate, a learned human-preference proxy our objective never optimizes. The recipe carries to text-to-image: applied to FLUX.2 [klein] \citep{flux2}, a 4B four-step generator, \method{} post-trains it into a one-step model that surpasses the four-step version on GenEval, 0.826 to 0.794, and on PickScore, \cvJoint{} to 22.58, in 90 H200 GPU-hours.
We summarize our contributions as follows.
\begin{itemize}
\item \textbf{A unifying framework.} We formalize distribution matching into a single paradigm, RDM, that needs no online teacher, and identify the two design axes that govern it, how the distributions are compared and the representations they are compared in. This lets us trace the quality ceiling of a method to a specific design choice rather than its headline idea.
\item \textbf{A simple recipe at the state of the art.} Varying each axis in isolation, we establish what drives quality: the right way to estimate the MMD, an exact within-batch repulsion paired with a Nystr\"om attraction to a frozen full-data reference; large fresh generation batches; a joint image-text objective on text-to-image tasks; and a constrained optimization that keeps a diverse encoder battery in balance. These choices combine into \method{}, which reaches state-of-the-art one-step ImageNet generation at \swmetric{} 1.30 against the real-data 1, with no online teacher, adversary, or reward model; the same recipe post-trains four-step FLUX.2 [klein] into a one-step model at a higher GenEval than the four-step base.
\item \textbf{A metric that resists gaming.} We evaluate with \swmetric{}, a Sliced-Wasserstein distance averaged over 14 encoders and real data scoring 1 by construction, an optimal-transport metric independent of the training loss and far harder to game than any single-encoder score.
\end{itemize}

\begin{figure}[t]
\centering
\includegraphics[width=0.8\linewidth]{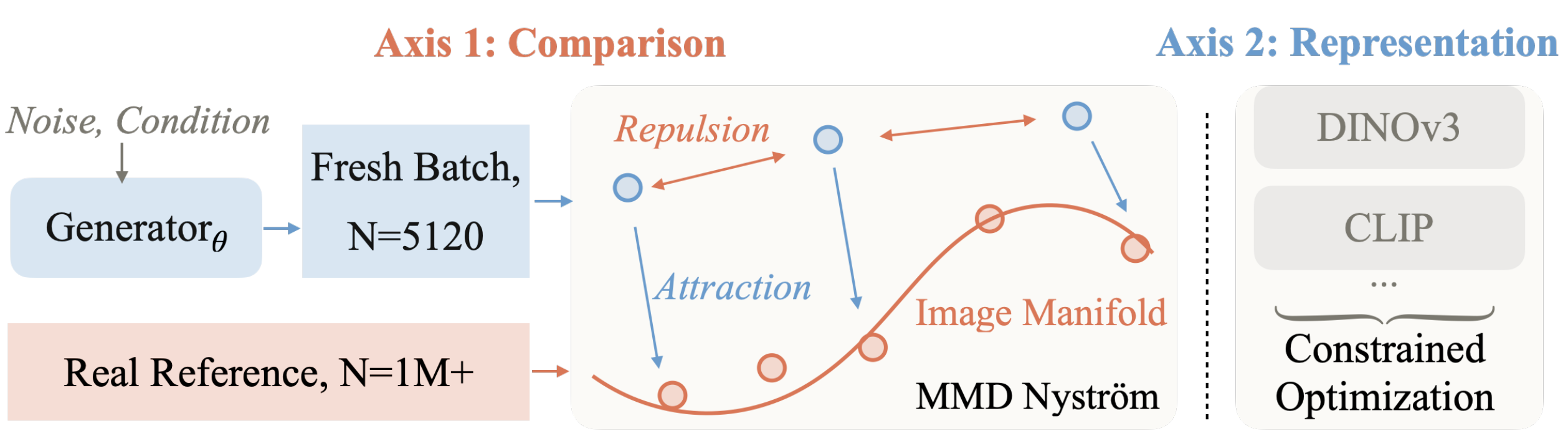}
\caption{\method{} trains a one-step generator by representation distribution matching alone: no online teacher, no adversary, no trajectory. Each step draws a fresh batch of $N$ samples and embeds it, together with a reference computed once and frozen, under a battery of ten pretrained encoders. In every feature space, generated samples are pulled toward the reference manifold by a Nystr\"om attraction and kept apart by an exact within-batch repulsion (\cref{eq:estimator}).}
\label{fig:method}
\end{figure}

\section{Related Work}

\paragraph{One-step and few-step generation.}
Diffusion and flow models \citep{ddpm,scoresde,flowmatching,edm,ldm,dit} pay an inference cost per denoising step. Step reduction either distills a pretrained teacher or removes it. Distillation matches the teacher's trajectory, score, or moments \citep{progdist,rectflow,diffinstruct,dmd,sid,mmdistill} or trains against an adversary \citep{add,dmd2}; the teacher-free route constrains the model on its own outputs or trajectories \citep{cm,ict,ect,scm,shortcut,meanflow,pmf,imf,imm}. RDM needs no online teacher and constrains no trajectory: it compares generated samples against a frozen reference directly.

\paragraph{Matching distributions in fixed feature spaces.}
Casting generation as distribution matching is the GAN program \citep{gan,improvedgan}; with fixed kernels it gave moment matching networks \citep{gmmn,dziugaite}, adversarial kernels \citep{mmdgan,kid}, and sliced Wasserstein generators \citep{swg,swgm}. What changed since is the feature space: frozen pretrained encoders, used for perceptual losses \citep{perceptual,lpips}, discriminator features \citep{projectedgan,visionaided}, and alignment targets \citep{repa}, now support direct feature-distribution matching, as the drifting field \citep{drifting}, the Fr\'echet-distance loss \citep{fdloss}, and a concurrent Sinkhorn flow \citep{wflow} show. The principle runs implicitly through this lineage; our contribution is to name it, chart its two design axes, and locate prior methods within them.

\section{Representation Distribution Matching and its design space}
\label{sec:framework}

A one-step generator $g_\theta$ maps a prior $z \sim p_z$ to an image in a single evaluation, with output law $p_\theta$. Given a frozen encoder $\phi$ that sends an image to a feature $\phi(x) \in \mathbb{R}^{D}$, RDM aligns the feature laws of generated and real data (\cref{fig:method}),
\begin{equation}
\mathcal{L}(\theta) = \mathcal{D}\!\left(\phi_* p_\theta,\; \phi_* p_{\mathrm{data}}\right),
\label{eq:rdm}
\end{equation}
where $\phi_*$ is the pushforward and $\mathcal{D}$ a distance between distributions. Constraining the output distribution rather than a per-sample trajectory makes the generator one-step by construction; the same objective post-trains a few-step sampler by treating its final output as $g_\theta$.

Every instance of \cref{eq:rdm} is fixed by two choices, the axes of this paper: \textbf{the comparison}, set by which discrepancy $\mathcal{D}$ scores the feature laws, which estimator computes it from finite samples, what reference stands in for each side, and which joint law is matched under conditioning (Sections~\ref{sec:estimator} and~\ref{sec:modelling}); and \textbf{the representations}, which encoders define the feature spaces and how several are weighted (Section~\ref{sec:multirep}).

Our decomposition locates prior methods on these axes and attributes each method's ceiling to a specific choice. The Fr\'echet-distance loss freezes a global data-side reference, the right call, but compresses it to two moments, so matching can saturate while images stay flawed. The drifting field has a sharp pairwise estimator, but it rebuilds its reference from every batch at a cost that confines it to small batches, exactly where a distribution estimate is noisiest. Both train against a few encoders under fixed weights, which Section~\ref{sec:multirep} shows is gameable. \method{} is the combination of the preferred choice on each axis.

\subsection{The comparison axis: choosing and estimating the discrepancy}\label{sec:estimator}

A positive definite kernel $k$ on feature space defines
\begin{equation}
\mathrm{MMD}^2(P, Q) \;=\; \mathbb{E}_{x,x' \sim P}\,k(x,x') \;-\; 2\,\mathbb{E}_{x \sim P,\, y \sim Q}\,k(x,y) \;+\; \mathbb{E}_{y,y' \sim Q}\,k(y,y'),
\label{eq:mmd}
\end{equation}
which vanishes exactly when $P = Q$ for a characteristic kernel such as the Gaussian \citep{gretton,sriperumbudur}. We adopt the squared MMD with this Gaussian kernel, $k(x,y) = \exp\!\big(\!-\lVert x - y\rVert_2^2 / 2\sigma_\phi^2\big)$, on the raw encoder embeddings; the bandwidth $\sigma_\phi$ is fixed per encoder by the median heuristic and held at a single scale. What a generator optimizes is a finite-sample estimate, and the estimator sets its cost, its variance, and the blind spots it can exploit.

Write $g_i = \phi(g_\theta(z_i))$ for the features of a generated batch of size $B$. Of the three terms of \cref{eq:mmd}, the data term is constant in $\theta$ and dropped; the cross term attracts generated features toward the data; the generator term repels them from one another, the only force preventing collapse onto the densest modes. The two demands are opposite, so we estimate the terms differently,
\begin{equation}
\widehat{\mathcal{L}}_\phi \;=\; \underbrace{\frac{1}{B^2} \sum_{i,j} k(g_i, g_j)}_{\text{repulsion, exact}} \;-\; \underbrace{\frac{2}{B} \sum_{i} \psi(g_i)^\top \bar\mu_\phi}_{\text{attraction, Nystr\"om}},
\label{eq:estimator}
\end{equation}
where $\psi$ is the Nystr\"om feature map and $\bar\mu_\phi$ the frozen reference mean embedding it induces over the full training set, both made precise below. Every batch is scored by all encoders in the battery, and we sum $\widehat{\mathcal{L}}_\phi$ over them each step with the adaptive weights of Section~\ref{sec:multirep}.

\paragraph{An exact repulsion, a frozen attraction.}
The two terms sum over different sets and we estimate them differently. The repulsion runs only within the batch, where the exact $B \times B$ kernel sum is cheap, so we leave it exact, one matrix per encoder. The attraction instead compares against the full training set: resampling it each step, as the standard two-sample estimator does, injects reference noise that grows as the bandwidth shrinks, so we compute it once and freeze it through a Nystr\"om kernel mean embedding \citep{nystromkme}. With $m{=}4096$ landmarks $\ell_j$ placed by $k$-means on the data features and kernel matrix $K_{mm}$, $\psi(x) = K_{mm}^{-1/2}\big(k(x, \ell_1), \ldots, k(x, \ell_m)\big)^\top$ makes $\psi(x)^\top \psi(y)$ the Nystr\"om approximation of $k(x, y)$, and $\bar\mu_\phi = \frac{1}{n} \sum_{t} \psi(r_t)$ is precomputed once over all $n = 1.28$M training images and frozen. Each step pulls the batch toward this zero-variance summary at cost $\mathcal{O}(Bm)$, negligible next to the encoder forward passes.

\paragraph{Why MMD, and why Nystr\"om.}
Each alternative discrepancy gives up one of these advantages: the Fr\'echet distance collapses each side to two moments and can saturate while samples stay flawed; sliced-Wasserstein relies on sorting within each projection, so it scores the batch only against a resampled batch rather than the full real distribution \citep{swg}; and the drifting field is a per-particle normalized form of the same MMD gradient, steadier at small batches but reducing to the plain MMD as the batch grows, its resampled per-batch reference keeping it small-batch \citep{drifting}. For the attraction term, Nystr\"om landmarks beat random Fourier features \citep{rff}: the landmark basis is data-dependent, centered on real points and accurate exactly where generation happens, whereas global cosines spend capacity over an ambient space the manifold barely occupies and leave unresolved directions that a generator under optimization pressure exploits. Theory concurs, with data-dependent bases dominating once the kernel spectrum decays quickly and $m$ of order $\sqrt{n}\,\log n$ landmarks retaining the exact embedding's $n^{-1/2}$ rate \citep{nysvsrff,nystromkme}. A controlled study makes both choices concrete.

\begin{figure}[t]
\centering
\includegraphics[width=0.8\linewidth]{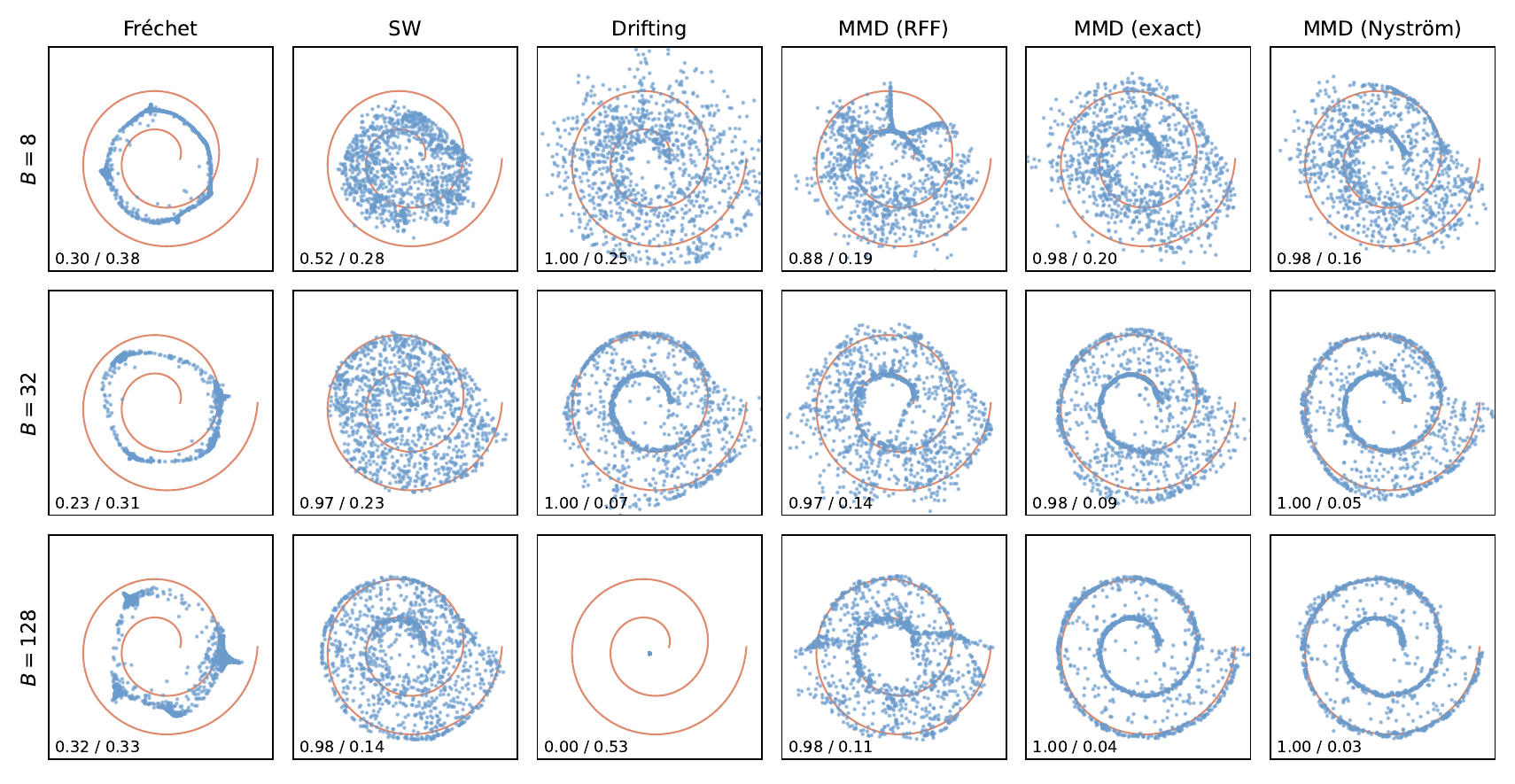}
\caption{Spiral diagnostic at ambient dimension $D = 64$. Rows sweep the training batch size $B$, and columns are different methods. Fr\'echet is the Gaussian 2-Wasserstein on a frozen global mean and covariance; sliced-Wasserstein uses $L = 1000$ resampled projections; drifting is the faithful coupled field, best-effort tuned with a reference bank of 128; and the MMD family uses a multi-scale Gaussian kernel with $m = 512$ features or landmarks per scale, where random features and Nystr\"om match a frozen global reference mean while the exact estimator sees $B$ reference samples per step. Corner numbers are anchor recall / medDist, the median distance to the curve, on which real data floors at 0.033. Fr\'echet is batch-insensitive and never traces the curve, its two moments cannot encode the manifold; the other distances fail by sampling instead, sliced-Wasserstein collapsing at small $B$, drifting at large $B$, and random features with dimension. Nystr\"om is the sharpest in every row and the only distance strong across all regimes.}
\label{fig:toy}
\end{figure}

\paragraph{A controlled study of the estimator.}
Real encoders place data on a thin manifold in a high-dimensional space, and we isolate this regime with a known target: following \citet{jit}, a two-turn spiral buried in $\mathbb{R}^{64}$ by a fixed orthonormal map, the same MLP generator trained under each objective at a matched budget while the batch sweeps $B \in \{8, 32, 128\}$, scored by anchor recall and medDist, the median distance to the curve, on which real data scores 0.033 (settings in \cref{fig:toy}).

At the largest batch both MMD estimators, MMD exact and MMD Nystr\"om, lock onto the spiral, while random features stay diffuse, sliced-Wasserstein stays loose, and drifting collapses. As the batch shrinks, MMD exact degrades as its per-batch reference thins, whereas MMD Nystr\"om pulls toward the same frozen reference at every batch size and stays sharpest in every row; sliced-Wasserstein loses recall at the smallest batch and drifting collapses at the largest. MMD Nystr\"om is the only method that fails nowhere.

\subsection{The comparison axis: batches and conditioning}\label{sec:modelling}

\paragraph{The generator side: large, fresh batches.}

\begin{wrapfigure}{r}{0.42\textwidth}
\centering
\includegraphics[width=0.4\textwidth]{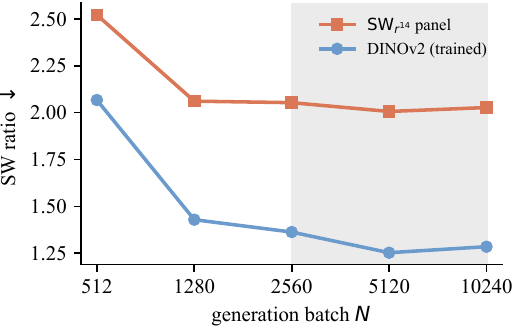}
\caption{Generation batch size $N$ at a matched wall-clock budget, fine-tuning a single-encoder DINOv2 Nystr\"om-MMD arm; Quality climbs with $N$ to a broad optimum (shaded).}
\label{fig:batch}
\end{wrapfigure}

With the data side frozen once over the full training set, the generated distribution is the only quantity still moving, and it moves at every step, so it must be sampled fresh; estimating it from a stale buffer, as the EMA queue of \citet{fdloss} does, biases the gradient off-policy. A fresh batch makes its size $N$ the operative variable: a larger $N$ lowers the variance of the estimate but, at a fixed compute budget, buys fewer optimizer steps, trading estimate sharpness against the number of updates. Large fresh batches are normally ruled out by memory, which gradient caching \citep{gradcache} removes by accumulating the exact full-batch gradient in chunks at the cost of one chunk. We sweep $N$ at a matched wall-clock budget, scaling the learning rate as $\sqrt{N}$ \citep{malladi} so every arm sees about one epoch split into more or fewer updates (\cref{fig:batch}). Quality climbs with $N$: the trained encoder sharpens while the held-out-dominated panel barely moves, the smallest batch is noise-dominated and regresses despite far more optimizer steps, and the curve then flattens into a broad optimum. We adopt $N{=}5120$ for ImageNet and the larger $N{=}10240$ for the FLUX post-training; exact values are in Appendix~\ref{app:batch}.
\paragraph{Conditional tasks: match the joint, not the marginal.}
\label{sec:joint}
A prompted generator can satisfy the image marginal while drifting from its prompts: realism bought with alignment. We instead match the joint law. With a frozen text encoder $\tau$ and coupled features $\Phi(x, c) = \phi(x) \oplus \tau(c)$,
\begin{equation}
\mathcal{L}_{\mathrm{joint}}(\theta) \;=\; \mathcal{D}\!\big(\Phi_* p_\theta,\; \Phi_* p_{\mathrm{data}}\big),
\label{eq:joint}
\end{equation}
where reference pairs couple each image with its caption and generated pairs couple each output with the prompt that produced it; the estimator is unchanged, landmarks now reference image-text pairs. Under the kernel a generated image is pulled toward reference images whose captions resemble its prompt, so prompt fidelity is part of what is matched. Post-training the four-step FLUX.2 [klein] \citep{flux2} into a one-step model with this objective surpasses the four-step version on GenEval \citep{geneval} while also surpassing its PickScore (Section~\ref{sec:t2i}); the marginal alternative sacrifices alignment with no compensating quality gain (Table~\ref{tab:geneval}).
\subsection{The representation axis: one encoder is never enough}
\label{sec:multirep}

Feature distances are also how realism is scored: FID and its descendants \citep{fid,kid,cmmd,fddinov2} reduce it to the distributional gap under one pretrained encoder, read as a proxy for human judgment. The proxy is fragile. FID falls under fringe ImageNet-class features with no gain in perceived quality \citep{fidclasses}, and such a distance is directly \emph{optimizable}: a generator can be driven below the score of real validation data while staying visibly fake \citep{fdloss}. The question this axis turns on: \emph{is there any feature space whose distance, once minimized, yields images humans cannot tell from real?}

\paragraph{Overfitting a single encoder.}

\begin{figure}[t]
\centering
\includegraphics[width=0.8\textwidth]{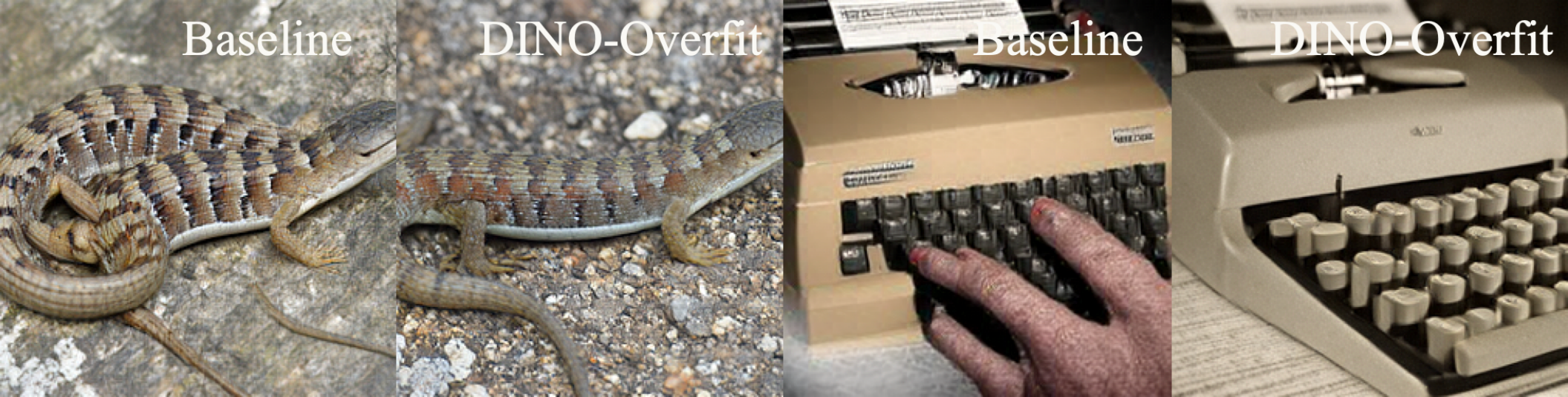}
\caption{Matching only DINOv2 features drives its distance to the real floor, $\mathrm{SW}_{\text{dino}}{=}1.01$, yet improves quality unevenly: the lizard (left) becomes indistinguishable from real, the typewriter (right) keeps clear artifacts. A saturated single-encoder distance does not imply realism.}
\label{fig:overfit}
\end{figure}

Below-real scores have so far been shown only on weak proxies, Inception and ConvNeXt, inviting the objection that a sufficiently rich encoder, once satisfied, would force realism. We test the hardest case we can construct: DINOv2, far more semantically structured, on which the base checkpoint starts far from real, $\mathrm{SW}_{\text{dino}}=1.81$. Matching it alone, $N{=}5120$ for $1000$ steps, drives the distance to $1.01$, essentially the real-validation floor of $1.00$: by DINOv2's account the generator is as close to real as real data. \cref{fig:overfit} says otherwise. The objective repairs some classes, the lizard becomes hard to tell from a photograph, and leaves others untouched, the typewriter keeps an implausible key layout at that same floor score. The limitation is single-encoder matching itself, not the choice of encoder, and the resolution is not a better encoder but a diverse ensemble.
\label{sec:overfit}

\paragraph{Constrained optimization against multiple encoders.}
\label{sec:battery}
A single encoder gives only a pseudometric, but the combined kernel of a diverse panel is characteristic and vanishes only at the real distribution \citep{gretton,sriperumbudur,mmdagg}; we therefore train against ten of the fourteen panel encoders (Appendix~\ref{app:encoders}), frozen backbones chosen to fail in different ways. The weighting then decides whether this diversity survives: under fixed weights the optimizer drives the aggregate down through whichever encoders are easiest. We instead pose the weighting as a constrained optimization, each encoder required to reach its real-validation floor with its weight the Lagrange multiplier, set by proportional control under a satisfaction gate, the proportional term of the PID-Lagrangian scheme of \citet{pidlag}. An encoder's excess $e_\phi = s_\phi - b_\phi$ sets its weight: those at or below their floor drop out, while the violators share a fixed budget through a softmax, $\lambda_\phi \propto \exp\!\big(e_\phi / (\tau\,\bar e)\big)$, so the representations farthest from real are weighted most; when all are satisfied the weights vanish, a natural anti-overfitting terminal state.

\paragraph{Scaling the multi-representation metric.}
\label{sec:metric}
Evaluation needs the same protection and must not collapse into the training loss. \citet{fdloss} aggregate a per-encoder ratio over a panel, the Fr\'echet form $\mathrm{FD}_{r^{k}}$; we keep that construction but replace the Fr\'echet distance with the Sliced-Wasserstein \citep{swg,swgm}, a proper optimal-transport distance that shares no estimator with the MMD we train against \citep{mind}. Our metric \swmetric{} averages the per-encoder ratio over the $k$ encoders,
\begin{equation}
\mathrm{SW}_{r^{k}} \;=\; \frac{1}{k}\sum_{e=1}^{k} r_e,
\qquad
r_e \;=\;
\frac{\mathrm{SW}\!\big(\phi_{e*}p_\theta,\;\phi_{e*}p_{\mathrm{train}}\big)}
     {\mathrm{SW}\!\big(\phi_{e*}p_{\mathrm{val}},\;\phi_{e*}p_{\mathrm{train}}\big)} .
\end{equation}
With $k=14$, real validation data scores $1$ by construction, a floor no released generator approaches (Table~\ref{tab:sw14}); four of the encoders are held out from training as a generalization check. Appendix~\ref{app:mmd} gives a kernel-MMD counterpart \mmmetric{}, and Section~\ref{sec:pickscore} validates \swmetric{} against PickScore.

\paragraph{Putting it together: \method{}.}
Together these choices define \method{}: an exact within-batch repulsion with a Nystr\"om attraction to a reference frozen once over the full data, large fresh generation batches, the joint image-text law on conditional tasks, and a diverse encoder battery balanced by constrained optimization. The reference $\bar\mu_\phi$ is precomputed once per encoder; each step then draws a fresh batch, generates in a single evaluation, encodes it under the training encoders with gradient caching, and sums the per-encoder losses of \cref{eq:estimator} under the proportional Lagrangian weights. Nothing else enters the objective: no online teacher, no adversary, no trajectory.

\section{Experiments}
\label{sec:experiments}

Sections~\ref{sec:estimator} to~\ref{sec:multirep} fixed each design choice with a controlled study in place. The experiments report what remains: the main results, one-step ImageNet generation and text-to-image post-training, and the ablations the studies did not cover.

\subsection{One-step ImageNet generation}
\label{sec:main}
\paragraph{Setup.} On ImageNet-256 \citep{imagenet}, we post-train the released pMF-H FD-SIM checkpoint \citep{pmf,fdloss} for $4000$ steps at learning rate $1.6\times10^{-6}$ and batch size $N{=}5120$ over the ten training encoders of Appendix~\ref{app:encoders}, each a Gaussian kernel at its median-heuristic bandwidth whose attraction is taken against the full $1.28$M-image ImageNet training set, compressed once into a $4096$-landmark Nystr\"om reference. The ten encoders are kept in balance by the proportional Lagrangian controller of Section~\ref{sec:battery} with a satisfaction gate over a fixed budget $\Sigma{=}10$, each encoder's real floor computed on the ImageNet validation set. Evaluation uses two off-objective measures. \swmetric{} is the Sliced-Wasserstein ratio averaged over the $14$-encoder panel, four encoders held out from training, estimated from $16384$ samples per set with $M{=}1024$ projections. PickScore \citep{pickscore} is a learned human-preference model scored against the class prompt; against the pMF-H FD-SIM start we render $4000$ class-conditional latents under matched noise with both models and report the paired mean and win rate.
\paragraph{Distributional quality.} Table~\ref{tab:sw14} places released ImageNet-256 generators on \swmetric{}; none approaches the real floor of $1$, the strongest reaching about $2.05$. \method{} sets the state of the art at \swmetric{} $1.30$, below every released generator, and is the best entry on nine of the fourteen encoders and on the aggregate. It cedes five: Inception, ConvNeXt, and MAE to the FD-loss model, which scores below real there by gaming a single space, DreamSim to that same model by a hair, and the held-out FLUX VAE to MAR-H. Appendix~\ref{app:mmd} reports the same field under \mmmetric{}, a kernel-MMD panel, which broadly agrees, with some reordering among the mid-field models.

\begin{table*}[t]
\centering
\caption{\swmetric{}, our primary metric, across released ImageNet-256 generators. Per-encoder floor-normalized $\mathrm{SW}$ ratio ($\mathrm{SW}(\text{gen},\text{train})/\mathrm{SW}(\text{val},\text{train})$, the Sliced-Wasserstein; $\approx\!1$ matches a fresh real draw, lower\,$=$\,closer). $\mathrm{SW}$ is an optimal-transport distance sharing no machinery with the kernel MMD of the training loss, so it cannot be gamed by matching the loss. $\mathrm{SW}_{r^{14}}$ is the arithmetic mean over the 14 encoders (matching \textsc{mmdr}$_{14}$'s aggregate). \colorbox{gray!15}{Grey} rows are one-step (single-NFE) models; $^{\star}$\,marks an external representation encoder in training. $^{\dagger}$\,marks the four encoders held out from training, namely DINOv2, SigLIP\,(v1), RADIO, and FLUX; $\mathrm{SW}_{r^{4}}^{\dagger}$ is the same floor-normalized mean restricted to those four, a generalization check.}
\label{tab:sw14}
\resizebox{\textwidth}{!}{%
\begin{tabular}{lrrrrrrrrrrrrrrrr}
\toprule
Model & \rotatebox{45}{Inception} & \rotatebox{45}{ConvNeXt} & \rotatebox{45}{DINOv2$^{\dagger}$} & \rotatebox{45}{MAE} & \rotatebox{45}{SigLIP2} & \rotatebox{45}{CLIP} & \rotatebox{45}{DINOv3} & \rotatebox{45}{SigLIP\,(v1)$^{\dagger}$} & \rotatebox{45}{PE-Core} & \rotatebox{45}{RADIO$^{\dagger}$} & \rotatebox{45}{WebSSL} & \rotatebox{45}{AIMv2} & \rotatebox{45}{DreamSim} & \rotatebox{45}{FLUX$^{\dagger}$} & \rotatebox{45}{$\mathrm{SW}_{r^{14}}\downarrow$} & \rotatebox{45}{$\mathrm{SW}_{r^{4}}^{\dagger}\downarrow$} \\
\midrule
\textit{Validation baseline} & 1.00 & 1.00 & 1.00 & 1.00 & 1.00 & 1.00 & 1.00 & 1.00 & 1.00 & 1.00 & 1.00 & 1.00 & 1.00 & 1.00 & 1.00 & 1.00 \\
\midrule
\rowcolor{gray!15}Drifting-L$^{\star}$ & 0.97 & 1.61 & 6.12 & 3.20 & 8.84 & 5.83 & 18.2 & 7.12 & 7.89 & 5.31 & 5.70 & 8.45 & 2.69 & 1.07 & 5.93 & 4.91 \\
\rowcolor{gray!15}iMF-XL & 0.96 & 1.22 & 5.07 & 2.96 & 6.89 & 5.04 & 15.2 & 6.01 & 7.26 & 4.19 & 4.75 & 7.06 & 2.62 & 1.03 & 5.02 & 4.08 \\
Open-MAGVIT2-L & 1.57 & 1.39 & 4.87 & 2.93 & 6.33 & 4.94 & 6.59 & 5.33 & 5.95 & 4.50 & 4.66 & 7.05 & 2.70 & 1.41 & 4.30 & 4.03 \\
SiT-XL/2 & 1.29 & 1.12 & 4.53 & 2.68 & 6.24 & 4.75 & 9.94 & 5.21 & 6.36 & 3.87 & 4.09 & 6.43 & 2.32 & 0.91 & 4.27 & 3.63 \\
\rowcolor{gray!15}pMF-H (base) & 1.25 & 0.88 & 3.91 & 3.15 & 6.43 & 3.93 & 6.62 & 4.88 & 6.69 & 4.00 & 4.25 & 6.87 & 2.63 & 1.71 & 4.09 & 3.63 \\
DiT-XL/2 & 1.38 & 0.99 & 4.12 & 2.51 & 5.50 & 4.72 & 8.92 & 4.91 & 5.98 & 3.50 & 3.82 & 6.01 & 2.35 & 1.03 & 3.98 & 3.39 \\
VAR-d30 & 1.08 & 1.12 & 4.18 & 3.18 & 5.71 & 4.51 & 6.37 & 5.19 & 6.52 & 3.77 & 3.88 & 6.28 & 2.63 & 0.88 & 3.95 & 3.51 \\
JiT-H & 1.07 & 1.46 & 3.73 & 2.78 & 5.52 & 6.17 & 5.34 & 4.84 & 6.59 & 3.85 & 3.75 & 6.11 & 2.81 & 1.13 & 3.94 & 3.39 \\
MDTv2-XL/2 & 0.86 & 0.98 & 3.76 & 2.44 & 5.53 & 4.71 & 9.78 & 4.94 & 6.33 & 3.08 & 3.59 & 5.48 & 2.30 & 0.79 & 3.90 & 3.14 \\
MAR-H & 1.00 & 1.04 & 4.16 & 2.39 & 5.25 & 4.34 & 9.05 & 4.86 & 6.21 & 3.23 & 4.01 & 6.04 & 2.20 & \textbf{0.48} & 3.87 & 3.18 \\
DDT-XL/2$^{\star}$ & 0.83 & 0.96 & 3.74 & 2.36 & 5.29 & 4.60 & 9.07 & 4.68 & 6.18 & 3.06 & 3.49 & 5.44 & 2.15 & 0.97 & 3.77 & 3.11 \\
SiT-XL/2+REPA$^{\star}$ & 0.85 & 1.01 & 3.63 & 2.35 & 5.13 & 4.33 & 8.15 & 4.58 & 5.99 & 3.02 & 3.34 & 5.29 & 2.12 & 0.72 & 3.61 & 2.99 \\
REG-XL$^{\star}$ & 0.79 & 0.91 & 3.13 & 2.03 & 4.68 & 3.92 & 7.09 & 4.02 & 5.74 & 2.60 & 2.88 & 4.65 & 1.83 & 0.71 & 3.21 & 2.62 \\
LightningDiT-XL$^{\star}$ & 0.89 & 0.90 & 3.25 & 2.04 & 4.44 & 3.76 & 4.90 & 3.92 & 5.22 & 2.80 & 3.29 & 5.18 & 1.99 & 0.78 & 3.10 & 2.69 \\
RAE-XL$^{\star}$ & 0.75 & 1.30 & 2.38 & 2.11 & 2.74 & 3.52 & 2.76 & 2.80 & 4.51 & 2.39 & 2.31 & 3.88 & 1.51 & 1.13 & 2.43 & 2.18 \\
REPA-E SiT-XL/1$^{\star}$ & 0.75 & 1.00 & 2.79 & 1.86 & 3.41 & 2.83 & 3.30 & 3.07 & 3.89 & 2.04 & 2.41 & 4.13 & 1.47 & 0.66 & 2.40 & 2.14 \\
\rowcolor{gray!15}pMF-H (FD-SIM)$^{\star}$ & \textbf{0.67} & \textbf{0.67} & 1.81 & \textbf{0.60} & 1.86 & 2.69 & 2.33 & 2.63 & 4.76 & 2.14 & 2.31 & 3.68 & \textbf{1.24} & 1.35 & 2.05 & 1.98 \\
\midrule
\rowcolor{gray!15}\method{} (ours)$^{\star}$ & 1.27 & 0.98 & \textbf{1.35} & 0.83 & \textbf{1.30} & \textbf{1.02} & \textbf{1.11} & \textbf{1.90} & \textbf{1.22} & \textbf{1.56} & \textbf{1.55} & \textbf{1.44} & 1.32 & 1.36 & \textbf{1.30} & \textbf{1.54} \\
\bottomrule
\end{tabular}}
\end{table*}

\begin{figure}[t]
\centering
\includegraphics[width=0.9\textwidth]{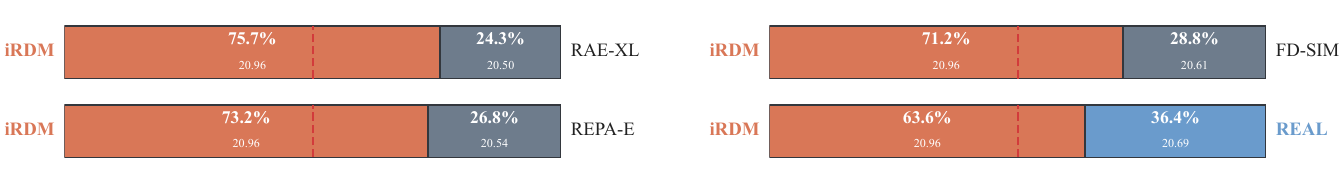}
\caption{PickScore preference, \method{} (orange) against prior generators and a real-photo reference; each bar shows the win rate, mean PickScore below. The FD-SIM bar is matched-noise paired, the others per-class means. \method{} is preferred over every prior generator and is, to our knowledge, the first one-step model to also surpass the held-out real-photo reference. The PickScore ordering agrees with the \swmetric{} ranking (Table~\ref{tab:sw14}), indicating that \swmetric{} also reflects human preference.}
\label{fig:pickscore_ext}
\end{figure}

\paragraph{Human preference.}\label{sec:pickscore} An off-objective check agrees. PickScore \citep{pickscore}, a learned human-preference model we never train against, prefers our converged checkpoint to its pMF-H FD-SIM start on $71.2\%$ of matched pairs ($20.61{\to}20.96$, paired $z{=}30.5$) and to the recent RAE-XL \citep{rae} and REPA-E SiT-XL \citep{repae} on $75.7\%$ and $73.2\%$ of classes (Figure~\ref{fig:pickscore_ext}); it even prefers our samples to held-out real photographs on $63.6\%$, to our knowledge the first one-step generator to pass the real-image PickScore.

\subsection{Text-to-image post-training}
\label{sec:t2i}

\paragraph{Setup.} We post-train FLUX.2 [klein] \citep{flux2} from its four-step checkpoint into a one-step model with the joint image-text objective of Section~\ref{sec:joint}, at batch size $N{=}10240$ and learning rate $2.83\times10^{-6}$ for $180$ steps, about $90$ H200 GPU-hours, under the encoder battery and constrained-optimization weighting of Section~\ref{sec:main}. The matching reference is collected once from the four-step teacher and then frozen, so the post-training queries no online teacher: a curated set of about $300$K teacher generations, PickScore-ranked COCO renderings \citep{coco} together with detector-verified GenEval-correct samples, compressed once into a Nystr\"om reference and detailed in Appendix~\ref{app:t2iref}. We evaluate with GenEval \citep{geneval} under its standard protocol and PickScore \citep{pickscore} on $500$ COCO validation prompts, and compare against a DMD2 \citep{dmd2} one-step distillation of the same four-step teacher (Appendix~\ref{app:dmd2}).

\paragraph{Results.} The one-step model surpasses its four-step start on GenEval overall, $0.826$ against $0.794$, with the per-category breakdown in Table~\ref{tab:geneval}: it matches the four-step version on single-object prompts, exceeds it on two-object, colors, position, and attribute binding, and trails only on counting. On PickScore it reaches \cvJoint{}, also above the four-step version's $22.58$. The DMD2 baseline reaches $0.804$ overall GenEval and $22.36$ PickScore, also listed in Table~\ref{tab:geneval}; Figure~\ref{fig:teaser}(c) traces both metrics over post-training compute.

\paragraph{Joint versus marginal.} The joint coupling carries the gain. A marginal variant that drops the caption from the feature, matching the image marginal alone with no SigLIP text concatenation, trails the joint model overall in Table~\ref{tab:geneval}, $0.801$ against $0.826$, and the gap concentrates on the categories that demand image-text alignment, two-object ($0.924$ against $0.899$) and attribute binding ($0.708$ against $0.608$), while single-object, which depends little on coupling, is essentially unchanged. Matching the joint law rather than the image marginal is what makes prompt fidelity part of the objective.

\begin{table}[t]
\centering
\caption{\textbf{GenEval and PickScore for one-step FLUX.2 [klein] post-training.} Per-category GenEval and PickScore of the four-step FLUX.2 [klein], the untrained one-step start, a DMD2 \citep{dmd2} baseline, the image-marginal ablation, and the joint one-step \method{}; best per column in bold. PickScore is scored on $500$ COCO validation prompts, higher is better. The joint image-text objective lifts the overall GenEval from $0.801$ (marginal) to $0.826$, surpassing the four-step version overall.}
\label{tab:geneval}
\resizebox{\textwidth}{!}{%
\begin{tabular}{lcccccccc}
\toprule
\textbf{Method} & \textbf{Single Obj.} & \textbf{Two Obj.} & \textbf{Counting} & \textbf{Colors} & \textbf{Position} & \textbf{Color Attr.} & \textbf{Overall} & \textbf{PickScore} \\
\midrule
FLUX.2 [klein] (4-step) & 0.994 & 0.904 & 0.791 & 0.880 & 0.575 & 0.623 & 0.794 & 22.58 \\
Untrained (1-step) & 0.894 & 0.323 & 0.603 & 0.673 & 0.225 & 0.128 & 0.474 & \cvUntr \\
DMD2 (1-step) & \textbf{0.997} & 0.894 & \textbf{0.806} & 0.864 & 0.603 & 0.660 & 0.804 & 22.36 \\
\method{} (1-step, marginal) & 0.991 & 0.899 & 0.763 & 0.910 & 0.638 & 0.608 & 0.801 & \cvMarg \\
\method{} (1-step) & 0.994 & \textbf{0.924} & 0.756 & \textbf{0.923} & \textbf{0.650} & \textbf{0.708} & \textbf{0.826} & \textbf{\cvJoint} \\
\bottomrule
\end{tabular}}
\end{table}

\FloatBarrier
\subsection{Constrained optimization versus uniform weighting}
\label{sec:ablation}
\paragraph{Setup.}

\begin{wraptable}{r}{0.46\textwidth}
\vspace{-0.8cm}
\centering
\caption{Per-encoder weighting: gated proportional Lagrangian versus uniform, $100$ steps from pMF-H on the \swmetric{} panel (lower is better, real floor $=1$). The gated controller edges uniform on the mean and clearly improves the worst encoder, the case the controller targets. Better arm in \textbf{bold}.}
\label{tab:pid}
\setlength{\tabcolsep}{5pt}
\begin{tabular}{lrrr}
\toprule
Aggregate & pMF-H & Gated & Uniform \\
\midrule
\swmetric{} & 2.09 & \textbf{1.88} & 1.90 \\
max  & 4.83 & \textbf{3.49} & 4.06 \\
\bottomrule
\end{tabular}
\end{wraptable}
We isolate the gated proportional Lagrangian controller of Section~\ref{sec:battery} against uniform weighting: both warm-start from pMF-H and train for $100$ steps under one recipe, with only the per-encoder allocation differing. The start is bimodal, the classic encoders already at or below their floor while the modern ones sit far from real, so the aggregate \swmetric{} of $2.09$ is set by the violators.

\paragraph{Results.} The gated controller pours the budget onto the worst encoder, PE-Core, while gating out the three already at their floor: it edges uniform on the mean, \swmetric{} $1.88$ against $1.90$, while decisively improving the worst case, $3.49$ against $4.06$ from a start of $4.83$, nearly twice the cut uniform manages (Table~\ref{tab:pid}). Reallocating toward the largest violation is what the controller targets; we make no claim here about perceptual quality, which this aggregate does not directly measure.

\subsection{The training distance}
\paragraph{Setup.} Holding the rest of the recipe fixed, we flip only the per-step distance across six fine-tuning losses, each warm-started from the same pMF-H checkpoint and fine-tuned against a single DINOv2 cls encoder for $100$ optimizer steps, sharing AdamW at learning rate $1.6\times10^{-6}$ and a generation batch of $5120$. The three kernel arms use an RBF kernel at bandwidth $\sigma{=}65$: \texttt{mmdx} is the biased $\mathrm{MMD}^2$ with an exact within-batch term and a Nystr\"om cross-term over $4096$ landmarks, \texttt{mmd\_exact} replaces that cross-term with the exact full generated-to-real pairwise mean, and \texttt{mmd\_rff} matches a frozen $4096$-dimensional random-Fourier-feature mean; \texttt{fd} is the Fr\'echet (Gaussian-moment) distance and \texttt{sw} a Sliced-Wasserstein loss with $128$ projections. The \texttt{drifting} arm is a faithful port of the published coupled force field across radii $\{0.2, 0.05, 0.02\}$, with in-batch generated negatives and real positives on the same features, run time-matched to the other arms (about $60$ steps at its generation batch of $8192$); we sweep its learning rate and report the gentlest, $1\times10^{-6}$, since its native $4\times10^{-4}$, tuned for from-scratch training, regresses the warm-start. We score every arm and the untrained baseline with two neutral third-party distances on the same features, a Sliced-Wasserstein ratio from $16384$ samples per set with $M{=}1024$ projections and an RFF-MMD ratio from $50000$ samples with $4096$ random Fourier features, so each arm is read on at least one distance it did not train on; the \texttt{sw} and \texttt{mmd\_rff} arms, which each optimize one of the two eval distances, are judged on the other (\cref{tab:lossfamily}).

\paragraph{Results.} One ranking holds across both, \texttt{mmdx} $\succ$ \texttt{mmd\_rff} $\succ$ \texttt{mmd\_exact} $\succ$ \texttt{fd} $\succ$ \texttt{sw} $\succ$ \texttt{drifting}, well above the untrained baseline: the three kernel-MMD estimators fill the top, moment matching follows, the Sliced-Wasserstein loss next, and a faithful port of the drifting force field is weakest even at the best of a learning-rate sweep. As an objective the Nystr\"om MMD moves the feature distribution closest to real, while optimal transport, an excellent judge, is among the least effective losses. Two controls confirm the reading: the exact full-pairwise MMD does not beat its Nystr\"om approximation, the low-rank cross-term being a smoother gradient, and training on a distance buys no advantage on that same distance, the Sliced-Wasserstein arm being beaten on the Sliced-Wasserstein eval by every kernel-MMD arm and even by moment matching. This is why \method{} trains with the MMD-Nystr\"om signal yet is evaluated with the independent Sliced-Wasserstein distance.

\begin{table}[t]
\centering
\caption{Training-distance ablation on DINOv2 (cls). The six fine-tuning losses warm-start the same pMF-H checkpoint and fine-tune against a single DINOv2 encoder, flipping only the per-step distance; \texttt{baseline} is pMF-H at step $0$. Each entry is a floor-normalized ratio (lower\,$=$\,closer to real, $\approx 1$ matches a fresh real draw) under two neutral distances, a Sliced-Wasserstein ratio (the per-encoder analogue of \swmetric{}) and an RFF-MMD ratio (that of \mmmetric{}). The order \texttt{mmdx}\,$\succ$\,\texttt{mmd\_rff}\,$\succ$\,\texttt{mmd\_exact}\,$\succ$\,\texttt{fd}\,$\succ$\,\texttt{sw}\,$\succ$\,\texttt{drifting} is identical on both; exact MMD does not beat Nystr\"om, and the SW-trained arm does not win the SW eval. \texttt{drifting} is a faithful port of the drifting force field shown at the best of a learning-rate sweep, its native rate regressing the warm-start.}
\label{tab:lossfamily}
\begin{tabular}{lccccccc}
\toprule
DINOv2 cls ratio ($\downarrow$) & baseline & mmdx & mmd\_rff & mmd\_exact & fd & sw & drifting \\
\midrule
SW & 1.927 & \textbf{1.420} & 1.466 & 1.492 & 1.547 & 1.583 & 1.746 \\
RFF-MMD & 10.393 & \textbf{4.495} & 4.839 & 5.438 & 5.798 & 6.413 & 8.258 \\
\bottomrule
\end{tabular}
\end{table}

\FloatBarrier

\section{Conclusion}

We have treated representation distribution matching, the principle behind a recent line of teacher-free one-step generators, as a design space rather than a collection of methods. Two axes fix every instance, how the generated and real feature distributions are compared and which representations they are compared in, and varying one at a time turns each into a preferred design with a mechanism behind it. On the comparison axis the classical MMD becomes a strong objective once estimated right, an exact within-batch repulsion paired with a Nystr\"om attraction toward a reference frozen once in advance, fed by large fresh generation batches and, on conditional tasks, by matching the joint image-text law rather than the image marginal. On the representation axis no single encoder is enough, since any one can be driven below the real score while samples stay visibly fake, so we match against a diverse battery of encoders held in balance by constrained optimization. Combining these choices gives \method{}, which sets the one-step state of the art on ImageNet at \swmetric{} $1.30$ and post-trains the four-step FLUX.2 [klein] into a one-step model that surpasses it on GenEval and PickScore, and we report the remaining gap with \swmetric{}, a Sliced-Wasserstein distance over the panel that shares no machinery with the training loss.

A gap to real remains: at \swmetric{} $1.30$ against a floor of $1$, the best one-step generator is still measurably short of a fresh real draw, and narrowing it is the natural next target. The design space leaves room to do so, through multi-scale kernels, learned or task-specific encoder panels, and richer conditional couplings, and the same recipe, a frozen reference matched by a single network evaluation, should transfer to modalities beyond images wherever a pretrained encoder supplies the feature space.

\section*{Acknowledgments}

We thank Jiawei Yang for helpful discussions. The project was partially funded by Valeo.

\bibliography{iclr2026_conference}
\bibliographystyle{plainnat}

\appendix
\section{Extended related work}
\label{app:related}

\paragraph{Scalable kernel estimators.}
Random Fourier features linearize kernel sums with data-independent bases \citep{rff}; Nystr\"om methods use data-dependent landmarks instead and dominate whenever the kernel spectrum decays quickly \citep{nysvsrff}. For the kernel mean embedding, the single point in the RKHS that summarizes a distribution \citep{kmesurvey}, Nystr\"om compression retains the full estimation rate of the exact embedding with far fewer landmarks than data points \citep{nystromkme}. Concurrently, DriftXpress accelerates drifting models by projecting the kernel field onto a low-rank RKHS with landmark approximations \citep{driftxpress}. Our use differs in target and in symmetry: DriftXpress approximates the full per-batch drifting update for speed, while we compress only the stationary data side, into a frozen global attraction target over 1.28M images, to remove reference noise, and keep the moving within-batch repulsion exact.

\paragraph{Metric gaming and multi-encoder evaluation.}
Single-encoder distances such as FID \citep{fid}, KID \citep{kid}, CMMD \citep{cmmd}, and feature-space precision and recall \citep{prdc} inherit the blind spots of their encoder: the scores move with resizing details \citep{cleanfid}, fall through fringe ImageNet-class features with no quality gain \citep{fidclasses}, and re-rank models when the encoder is swapped \citep{fddinov2}. \citet{fdloss} push this to the limit, driving a trained generator below the score of the real validation set; we show the failure is single-encoder matching itself rather than any weak encoder. When the proxy is learned, the same phenomenon is studied as reward hacking \citep{goodhart,rewardhacking,rmoveropt}, ensembling the proxies mitigates it \citep{rmensemble}, and Lagrangian methods control it in constrained reinforcement learning \citep{pidlag}. In our setting every proxy is frozen, so gaming pressure concentrates on the encoder weighting, which is exactly what the proportional Lagrangian controller regulates; evaluation aggregates 14 encoders, 4 held out from training, into \swmetric{}, a Sliced-Wasserstein distance independent of the loss.

\paragraph{Post-training text-to-image models.}
Few-step text-to-image systems are typically distilled from a multi-step teacher, adversarially \citep{add} or by score-based distribution matching \citep{dmd2}, and then steered toward human taste by optimizing learned preference rewards. The reward models are trained from human choices \citep{imagereward,pickscore} and optimized either by policy gradients \citep{ddpo} or by direct preference objectives \citep{diffusiondpo}; all inherit the gameability of a single learned scorer, the same axis our multi-encoder battery addresses for frozen proxies. For our text-to-image result the four-step FLUX.2 [klein] \citep{flux2} generates a reference set in advance, and the joint image-text objective of Section~\ref{sec:joint} matches the one-step model against it with no online teacher and no reward model, evaluated by GenEval \citep{geneval} and PickScore.

\paragraph{The evaluation landscape.}
The generators placed on \swmetric{} in Table~\ref{tab:sw14} span the current families: latent diffusion transformers \citep{dit,sit,mdt,ddt,lightningdit}, representation-aligned variants that inject external encoder features during training \citep{repa,reg,rae}, autoregressive and masked-token models \citep{var,mar,openmagvit2}, pixel-space transformers \citep{jit}, and the one-step MeanFlow, drifting, and FD-loss lines \citep{pmf,imf,drifting,fdloss}. The models that use an external representation encoder in training, the starred rows of the table, populate its strongest entries, consistent with the premise that representation supervision is the operative ingredient; RDM makes that ingredient explicit and studies it in isolation.

\section{Encoder panel}
\label{app:encoders}

\mmmetric{} takes the arithmetic mean over the $14$ encoders of \cref{tab:encoders}, each frozen at its released weights and read out as a single pooled image embedding $\phi(x)$ at the listed input resolution, with no feature normalization. The panel deliberately spans training paradigms, supervised classification, self-supervised distillation and masked reconstruction, language supervision, multi-teacher agglomeration, multimodal autoregression, human similarity tuning, and a generative autoencoder, so the representations fail in different ways. Ten supervise training; the four held out for evaluation only are DINOv2, SigLIP\,(v1), C-RADIOv3-L, and the FLUX VAE.

\begin{table*}[t]
\centering
\caption{The fourteen-encoder panel. Each backbone is frozen at its released weights and $\phi(x)$ is its pooled image embedding, taken at the listed input resolution. Pool: \textsc{cls} class token, \textsc{avg} mean over patch or spatial tokens, \textsc{attn} attention-pooling head. Ten encoders supervise training; four are held out for evaluation only.}
\label{tab:encoders}
\resizebox{\textwidth}{!}{%
\begin{tabular}{lllccr}
\toprule
Encoder & Checkpoint & Architecture & Input & Pool & $D$ \\
\midrule
\multicolumn{6}{l}{\textit{Training panel (10)}}\\
Inception-v3 \citep{inceptionv3} & FID Inception-v3 & CNN & $299$ & avg & $2048$ \\
ConvNeXt\,V2-B \citep{convnext} & \texttt{convnextv2\_base.fcmae\_ft\_in22k\_in1k} & CNN & $224$ & avg & $1024$ \\
MAE \citep{mae} & \texttt{vit\_large\_patch16\_224.mae} & ViT-L/16 & $224$ & avg & $1024$ \\
CLIP \citep{clip} & \texttt{vit\_large\_patch14\_clip\_224.openai} & ViT-L/14 & $256$ & cls & $1024$ \\
DINOv3-L \citep{dinov3} & \texttt{vit\_large\_patch16\_dinov3.lvd1689m} & ViT-L/16 & $224$ & cls & $1024$ \\
PE-Core-L \citep{pecore} & \texttt{vit\_pe\_core\_large\_patch14\_336.fb} & ViT-L/14 & $224$ & attn & $1024$ \\
SigLIP2-So400m \citep{siglip2} & \texttt{vit\_so400m\_patch16\_siglip\_256.v2\_webli} & ViT-So400m/16 & $224$ & attn & $1152$ \\
AIMv2-H \citep{aimv2} & \texttt{aimv2\_huge\_patch14\_224.apple\_pt} & ViT-H/14 & $224$ & avg & $1536$ \\
Web-SSL DINO 1B \citep{webssl} & \texttt{webssl-dino1b-full2b-224} & ViT-1B & $224$ & cls & $1536$ \\
DreamSim \citep{dreamsim} & DINO\,+\,CLIP\,+\,OpenCLIP ensemble & ViT ens. & $224$ & cls & $1792$ \\
\midrule
\multicolumn{6}{l}{\textit{Held out (4)}}\\
DINOv2 \citep{dinov2} & \texttt{vit\_large\_patch14\_dinov2.lvd142m} & ViT-L/14 & $256$ & cls & $1024$ \\
SigLIP\,(v1) \citep{siglip} & \texttt{vit\_so400m\_patch14\_siglip\_384.webli} & ViT-So400m/14 & $384$ & attn & $1152$ \\
C-RADIOv3-L \citep{amradio,cradio} & NVIDIA C-RADIOv3-L & ViT-L, multi-teacher & $256$ & summary & $3072$ \\
FLUX VAE \citep{fluxvae} & FLUX.1 VAE, $4{\times}4$ patch-mean & VAE & $256$ & patch-mean & $1024$ \\
\bottomrule
\end{tabular}}
\end{table*}

\section{Batch-size sweep}
\label{app:batch}

Table~\ref{tab:batch} tabulates the sweep plotted in \cref{fig:batch}.

\begin{table}[t]
\centering
\caption{Generation batch size $N$ at a matched wall-clock budget ($\approx 6000$\,s each), fine-tuning a single-encoder DINOv2 Nystr\"om-MMD arm; entries are Sliced-Wasserstein ratios (lower is closer to real). The smallest batch regresses above the untrained base despite the most optimizer steps; the optimum is broad, with $N{=}10240$ only marginally worse than $N{=}5120$.}
\label{tab:batch}
\begin{tabular}{lccc}
\toprule
Batch $N$ & lr & DINOv2 $\downarrow$ & $\mathrm{SW}_{r^{14}}\downarrow$ \\
\midrule
\textit{untrained base} & n/a & $1.927$ & $2.085$ \\
\midrule
$512$    & $5.1{\times}10^{-7}$ & $2.067$ & $2.521$ \\
$1280$   & $8.0{\times}10^{-7}$ & $1.429$ & $2.061$ \\
$2560$   & $1.1{\times}10^{-6}$ & $1.363$ & $2.053$ \\
$5120$   & $1.6{\times}10^{-6}$ & $\mathbf{1.253}$ & $\mathbf{2.006}$ \\
$10240$  & $2.3{\times}10^{-6}$ & $1.285$ & $2.027$ \\
\bottomrule
\end{tabular}
\end{table}

\section{Kernel-MMD evaluation}
\label{app:mmd}

Table~\ref{tab:mmdr14} reports \mmmetric{}, the training-aligned kernel-MMD cross-check of our primary \swmetric{} metric, over the full released field on the $14$-encoder panel. Each entry is a per-encoder RFF-MMD ratio against real training data, with real validation scoring $1$ by construction, and \mmmetric{} is their arithmetic mean over the 14 encoders. The ordering broadly agrees with \swmetric{} (Table~\ref{tab:sw14}), with some reordering among the mid-field models; because the loss is itself a kernel MMD, a single encoder can be pushed below the real floor here, Inception at $0.22$ for the FD-SIM model, which the optimal-transport \swmetric{} and the held-out split resist.

\begin{table*}[t]
\centering
\caption{MMD-RFF distance ratio (\textsc{mmdr}; lower\,$=$\,closer to real) of released ImageNet-256 generators and our \method{} across 14 vision encoders. $\overline{\textsc{mmdr}}_{14}$ is the arithmetic mean over the 14 encoders. The \textit{validation baseline} is \textsc{mmdr}\,$=$\,1 by definition (real held-out data); parentheses give the raw $\mathrm{mmd}^2(\text{val},\text{train})\!\times\!10^{3}$ normaliser. \colorbox{gray!15}{Grey} rows are one-step (single-NFE) models. $^{\star}$\,marks an external representation encoder in training (REPA/RAE-style alignment, FD-loss, or drift-loss on encoder features). Strongest at bottom.}
\label{tab:mmdr14}
\setlength{\tabcolsep}{3pt}
\resizebox{\textwidth}{!}{%
\begin{tabular}{lrrrrrrrrrrrrrrr}
\toprule
Model & \rotatebox{45}{Inception} & \rotatebox{45}{ConvNeXt} & \rotatebox{45}{DINOv2} & \rotatebox{45}{MAE} & \rotatebox{45}{SigLIP2} & \rotatebox{45}{CLIP} & \rotatebox{45}{DINOv3} & \rotatebox{45}{SigLIP} & \rotatebox{45}{PE-Core} & \rotatebox{45}{RADIO} & \rotatebox{45}{WebSSL} & \rotatebox{45}{AIMv2} & \rotatebox{45}{DreamSim} & \rotatebox{45}{FLUX} & \rotatebox{45}{$\overline{\textsc{mmdr}}_{14}\downarrow$} \\
\midrule
\textit{Validation baseline} & \makecell[r]{1.00\\{\scriptsize(0.321)}} & \makecell[r]{1.00\\{\scriptsize(0.535)}} & \makecell[r]{1.00\\{\scriptsize(0.0455)}} & \makecell[r]{1.00\\{\scriptsize(0.787)}} & \makecell[r]{1.00\\{\scriptsize(0.103)}} & \makecell[r]{1.00\\{\scriptsize(0.600)}} & \makecell[r]{1.00\\{\scriptsize(0.0805)}} & \makecell[r]{1.00\\{\scriptsize(0.420)}} & \makecell[r]{1.00\\{\scriptsize(0.565)}} & \makecell[r]{1.00\\{\scriptsize(0.156)}} & \makecell[r]{1.00\\{\scriptsize(0.0363)}} & \makecell[r]{1.00\\{\scriptsize(0.181)}} & \makecell[r]{1.00\\{\scriptsize(0.209)}} & \makecell[r]{1.00\\{\scriptsize(0.346)}} & \makecell[r]{1.00\\{\scriptsize(0.313)}} \\
\midrule
\rowcolor{gray!15}Drifting-L$^{\star}$ \citep{drifting} & 0.80 & 3.61 & 136 & 21.1 & 157 & 53.0 & 845 & 64.2 & 92.0 & 52.5 & 133 & 128 & 11.8 & 3.98 & 122 \\
\rowcolor{gray!15}iMF-XL \citep{imf} & 0.87 & 2.08 & 91.0 & 17.7 & 98.1 & 40.2 & 594 & 46.4 & 79.0 & 35.0 & 92.9 & 93.1 & 10.8 & 3.20 & 86.1 \\
SiT-XL/2 \citep{sit} & 1.73 & 1.77 & 75.3 & 14.3 & 79.0 & 35.1 & 258 & 35.3 & 60.9 & 29.5 & 68.0 & 76.7 & 7.56 & 2.08 & 53.2 \\
Open-MAGVIT2-L \citep{openmagvit2} & 2.79 & 2.72 & 85.9 & 16.5 & 84.0 & 36.1 & 114 & 37.6 & 52.9 & 38.4 & 90.6 & 96.0 & 11.3 & 5.55 & 48.2 \\
MDTv2-XL/2 \citep{mdt} & 0.63 & 1.23 & 50.0 & 11.8 & 60.5 & 34.8 & 254 & 31.9 & 59.9 & 19.0 & 51.3 & 56.8 & 7.82 & 1.69 & 45.8 \\
MAR-H \citep{mar} & 0.79 & 1.19 & 61.5 & 11.0 & 56.5 & 28.7 & 219 & 30.1 & 57.4 & 20.7 & 65.0 & 68.1 & 7.18 & 0.37 & 44.8 \\
DiT-XL/2 \citep{dit} & 2.11 & 1.35 & 59.9 & 12.7 & 62.2 & 34.5 & 204 & 31.9 & 53.4 & 24.1 & 57.4 & 67.9 & 8.00 & 1.96 & 44.4 \\
\rowcolor{gray!15}pMF-H (base) \citep{pmf} & 1.78 & 0.91 & 54.6 & 17.8 & 87.5 & 22.4 & 115 & 30.5 & 65.7 & 31.5 & 70.6 & 94.4 & 10.9 & 8.91 & 43.7 \\
DDT-XL/2$^{\star}$ \citep{ddt} & 0.46 & 1.20 & 49.6 & 11.1 & 57.1 & 33.0 & 213 & 28.5 & 57.2 & 18.3 & 48.8 & 55.2 & 6.61 & 1.55 & 41.5 \\
VAR-d30 \citep{var} & 1.13 & 1.73 & 63.8 & 18.7 & 67.4 & 30.3 & 108 & 34.7 & 61.1 & 29.7 & 62.4 & 75.9 & 11.4 & 1.34 & 40.6 \\
JiT-H \citep{jit} & 1.26 & 3.06 & 48.1 & 14.2 & 66.0 & 51.1 & 74.4 & 31.6 & 64.1 & 30.4 & 60.7 & 73.3 & 12.3 & 2.56 & 38.1 \\
SiT-XL/2+REPA$^{\star}$ \citep{repa} & 0.56 & 1.37 & 47.2 & 11.1 & 55.4 & 29.3 & 171 & 27.7 & 54.1 & 18.4 & 45.3 & 52.5 & 6.40 & 1.37 & 37.3 \\
REG-XL$^{\star}$ \citep{reg} & 0.44 & 1.06 & 33.9 & 8.02 & 46.4 & 24.0 & 127 & 21.3 & 49.5 & 13.3 & 31.8 & 39.2 & 4.74 & 1.44 & 28.7 \\
LightningDiT-XL$^{\star}$ \citep{lightningdit} & 0.67 & 0.92 & 36.8 & 8.12 & 41.5 & 21.3 & 61.7 & 19.9 & 39.9 & 14.1 & 41.6 & 50.3 & 5.93 & 1.63 & 24.6 \\
REPA-E SiT-XL/1$^{\star}$ \citep{repae} & 0.34 & 1.26 & 26.1 & 6.19 & 24.0 & 11.9 & 26.9 & 12.9 & 21.9 & 7.97 & 21.6 & 31.8 & 2.64 & 0.87 & 14.0 \\
RAE-XL$^{\star}$ \citep{rae} & 0.36 & 2.34 & 19.0 & 7.36 & 14.9 & 16.6 & 18.4 & 10.4 & 28.8 & 11.1 & 18.5 & 26.7 & 3.32 & 4.12 & 13.0 \\
\rowcolor{gray!15}pMF-H (FD-SIM)$^{\star}$ \citep{fdloss} & 0.22 & 0.36 & 10.3 & 0.37 & 6.34 & 10.5 & 12.5 & 9.39 & 33.2 & 8.72 & 18.5 & 24.5 & 2.28 & 6.56 & 10.3 \\
\midrule
\rowcolor{gray!15}\method{} (ours)$^{\star}$ & 1.54 & 0.98 & 3.52 & 0.69 & 4.76 & 1.17 & 1.39 & 2.69 & 1.62 & 3.16 & 5.31 & 3.12 & 1.80 & 5.83 & \textbf{2.69} \\
\bottomrule
\end{tabular}}
\end{table*}

\section{Text-to-image post-training details}
\label{app:t2i}

\subsection{Reference curation}
\label{app:t2iref}
The text-to-image objective of Section~\ref{sec:t2i} matches the one-step model against a reference collected once from the four-step FLUX.2 [klein] teacher and then frozen, so the teacher is never queried during post-training. The reference concatenates two independently curated blocks of teacher generations, a perception block and a composition block, roughly $300$K image-caption pairs in all, each image kept with the caption that produced it for the joint kernel.

\paragraph{Perception block.} For each of the $82{,}783$ COCO train2014 images \citep{coco}, one caption apiece, the four-step teacher draws $24$ candidates, which PickScore \citep{pickscore} ranks; we keep the three highest per caption, giving $248{,}349$ pairs at full coverage. Selecting three of twenty-four oversampled draws anchors the reference on high-quality renderings of natural captions, supplying the perceptual side of the match.

\paragraph{Composition block.} To pull the model toward verified-correct composition rather than the teacher's average, which fails a large fraction of the harder compositional prompts, we keep only teacher generations a detector certifies as correct. For the $553$ GenEval \citep{geneval} prompts the teacher is sampled at $150$ seeds per prompt, topped up where a prompt has fewer than $100$ correct, and every generation is scored by the standard GenEval Mask2Former detector: a sample passes only when all prompt objects are present at the required count, color, and spatial relation. Capping at $100$ correct per prompt yields $53{,}800$ verified images covering $551$ of the $553$ prompts; two position prompts admit no correct teacher sample even at $1000$ seeds. Measured per seed over this pool, the teacher's correctness ranges from $98\%$ on single-object prompts down to $40\%$ on attribute binding and $33\%$ on position, so the filter most reshapes exactly the binding and spatial prompts on which the one-step model later improves.

\paragraph{Joint reference and prompt pool.} The two blocks are embedded under the ten training encoders of Appendix~\ref{app:encoders}, each image feature concatenated with its caption's frozen SigLIP2 text embedding \citep{siglip2} and compared under one Gaussian kernel at $0.25$ of the median-heuristic bandwidth with the text component weighted at $1$; the stack is compressed once into an $8192$-landmark Nystr\"om reference per encoder. The generator's conditioning pool mirrors the reference: the $82{,}783$ COCO captions together with the GenEval prompts replicated so the GenEval share of the pool matches that of the reference, about $18\%$. Aligning the generated and reference prompt distributions keeps the match well-posed, its optimum reached when the two coincide.

\subsection{DMD2 baseline}
\label{app:dmd2}
The DMD2 \citep{dmd2} baseline distills the same four-step FLUX.2 [klein] teacher into a one-step student. The released DMD2 targets an SD-UNet under $\epsilon$-prediction; we re-implement it for klein's flow-matching parameterization with the method intact: three networks, the one-step generator under training, a trainable critic estimating the student distribution's score, and the frozen four-step teacher, the DMD gradient being the teacher-minus-critic score difference, the critic updated every step and the generator every fifth. The student is initialized by regressing the teacher to a single step along its sampling ODE, after which distillation proceeds. We train at a global batch of $128$ at $512^2$ on four H200 GPUs with AdamW and no EMA.

\paragraph{Timestep schedule.} The one change klein's parameterization forces is the distillation timestep distribution. klein is guidance-distilled and its velocity is faithful only at high noise, the four native sampling nodes; drawing the timestep uniformly reaches a low-noise regime where the teacher collapses to the mode-averaged posterior mean and drives the generator below its initialization. Mapping the uniform draw through klein's signal-to-noise shift with the empirical $\mu \approx 2.03$ that matches the native node spacing, with a learning rate of $5\times10^{-7}$ and a short warmup, turns this divergence into a student that exceeds the teacher on GenEval.

\paragraph{Configuration and result.} Distillation quality is set mainly by the prompt distribution. Training on a broader LAION caption pool rather than COCO captions alone lifts GenEval at every checkpoint and softens the peak-then-erode profile of distribution-matching distillation, a $4.7$-point collapse becoming a $1.1$-point plateau. The reported baseline is the best configuration, the LAION-prompt run at its $500$-step peak, GenEval $0.804$ and PickScore $22.36$ (Table~\ref{tab:dmd2}); a variant adding a GAN term on real latents was metric-neutral, its discriminator separating real from generated latents so fast that the adversarial gradient vanished against the distribution-matching one, and is not reported. The student peaks in about $10$ H200 GPU-hours.

\begin{table}[t]
\centering
\caption{DMD2 \citep{dmd2} one-step student over distillation steps, best LAION-prompt configuration, against the four-step FLUX.2 [klein] teacher. GenEval under the standard protocol and PickScore on the $500$ COCO validation prompts; the $500$-step peak is the baseline reported in Table~\ref{tab:geneval}.}
\label{tab:dmd2}
\begin{tabular}{lcc}
\toprule
Step & GenEval & PickScore \\
\midrule
Teacher (4-step) & 0.794 & 22.58 \\
\midrule
$250$ & 0.778 & 22.17 \\
$500$ (reported) & \textbf{0.804} & 22.36 \\
$750$ & 0.792 & 22.36 \\
$1000$ & 0.793 & 22.27 \\
\bottomrule
\end{tabular}
\end{table}

\section{One-step text-to-image samples}
\label{app:t2isamples}

\begin{figure}[p]
\centering
\includegraphics[width=0.32\textwidth]{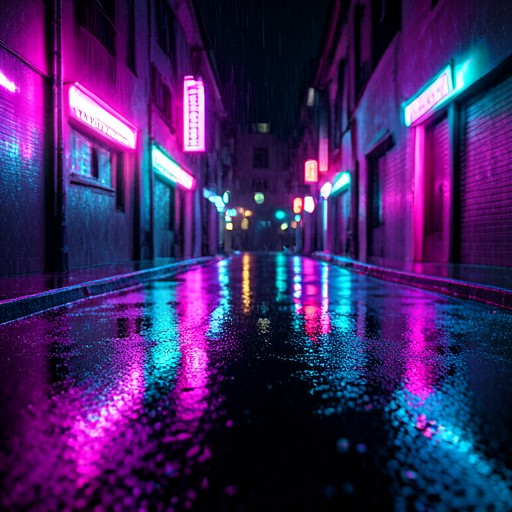}\hfill
\includegraphics[width=0.32\textwidth]{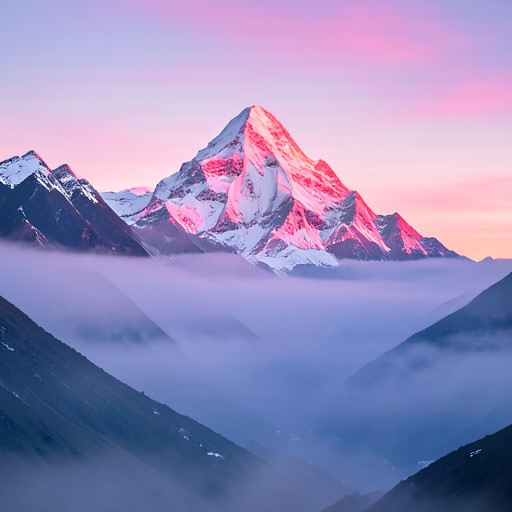}\hfill
\includegraphics[width=0.32\textwidth]{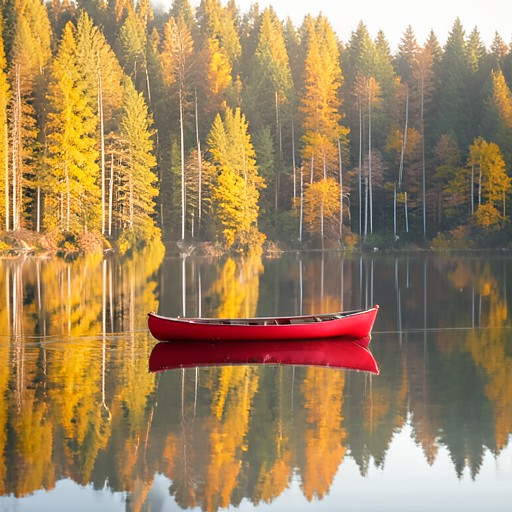}\\[3pt]
\includegraphics[width=0.32\textwidth]{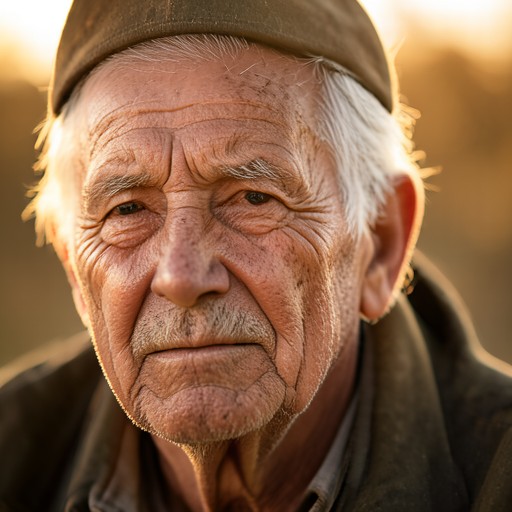}\hfill
\includegraphics[width=0.32\textwidth]{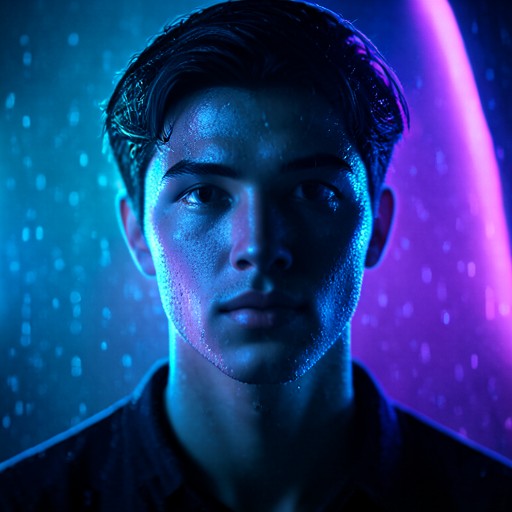}\hfill
\includegraphics[width=0.32\textwidth]{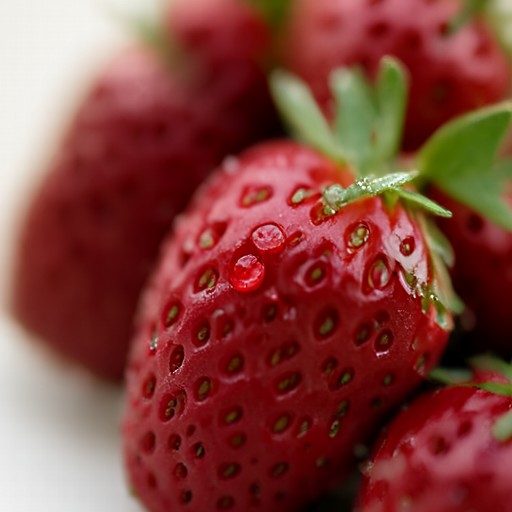}\\[3pt]
\includegraphics[width=0.32\textwidth]{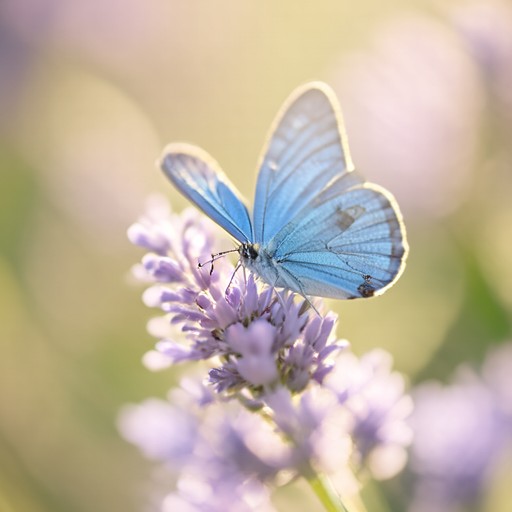}\hfill
\includegraphics[width=0.32\textwidth]{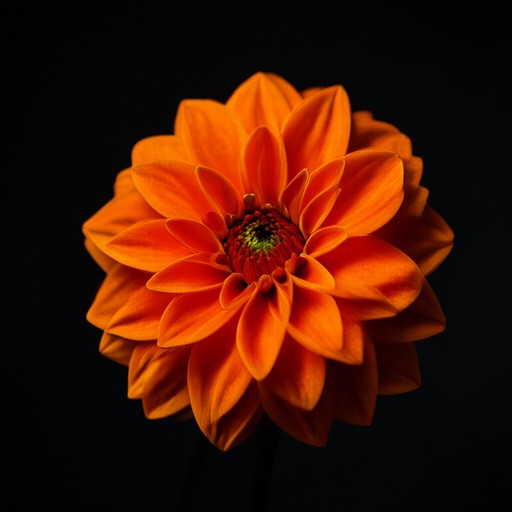}\hfill
\includegraphics[width=0.32\textwidth]{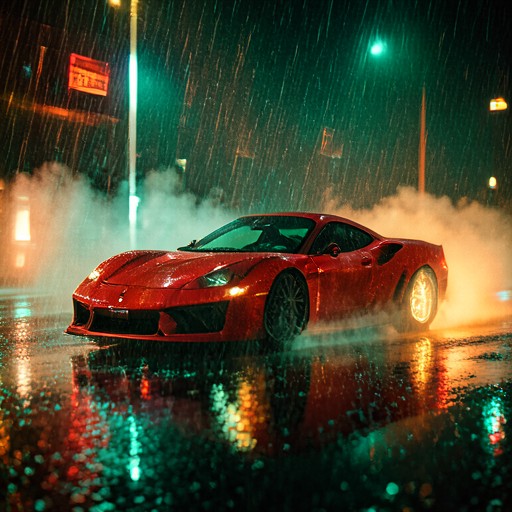}\\[3pt]
\includegraphics[width=0.32\textwidth]{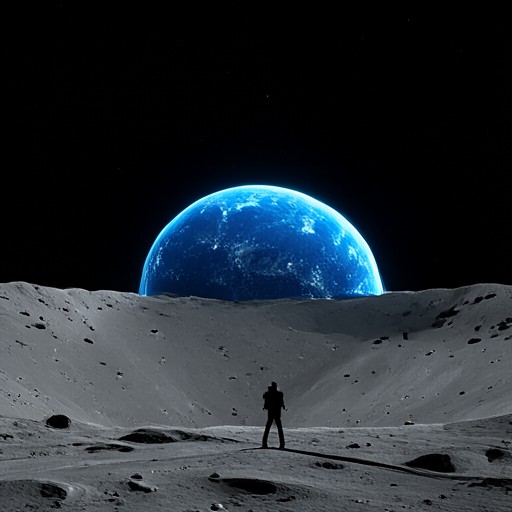}\hfill
\includegraphics[width=0.32\textwidth]{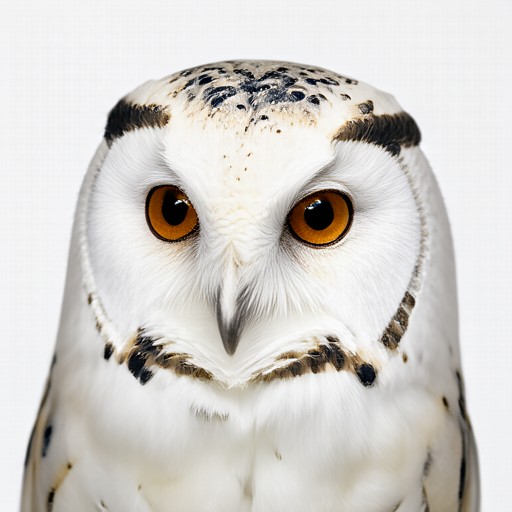}\hfill
\includegraphics[width=0.32\textwidth]{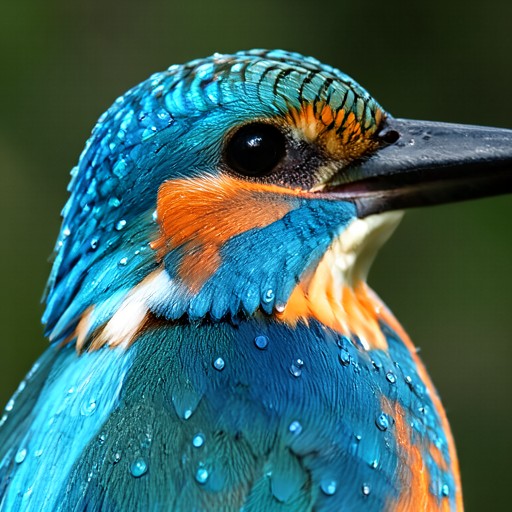}
\caption{\textbf{One-step text-to-image samples from \method{}.} Single-step generations from the post-trained one-step FLUX.2 [klein], $512\times512$, one network evaluation each.}
\label{fig:t2isamples}
\end{figure}

Figure~\ref{fig:t2isamples} shows additional single-step \method{} generations from the post-trained four-step FLUX.2 [klein], each a $512\times512$ image produced in one network evaluation.

\section{Qualitative comparison}
\label{app:qualitative}

\begin{figure}[t]
\centering
\makebox[0.5\textwidth][c]{\small\textbf{\textcolor[HTML]{C04A2B}{iRDM (ours)}}}%
\makebox[0.5\textwidth][c]{\small\textbf{pMF-H FD-SIM}}\\[2pt]
\includegraphics[width=\textwidth]{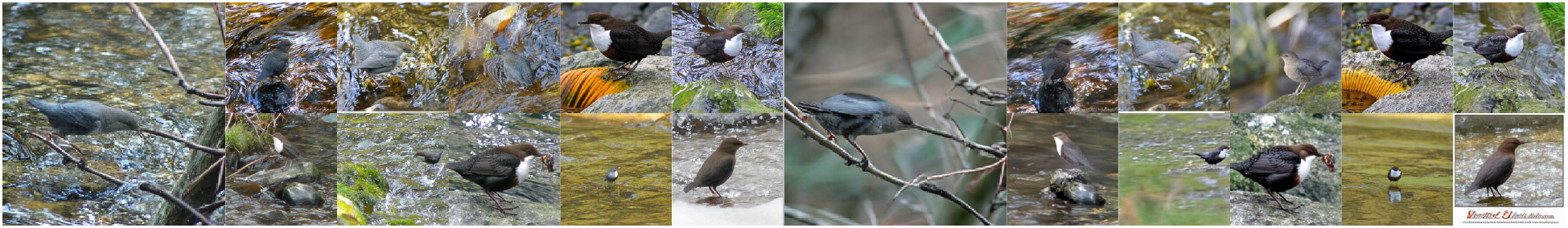}\\
{\scriptsize\itshape water ouzel}\\[3pt]
\includegraphics[width=\textwidth]{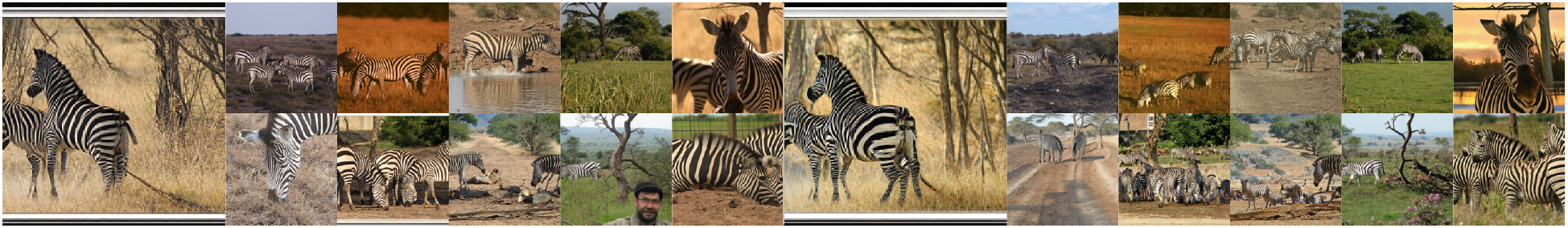}\\
{\scriptsize\itshape zebra}\\[3pt]
\includegraphics[width=\textwidth]{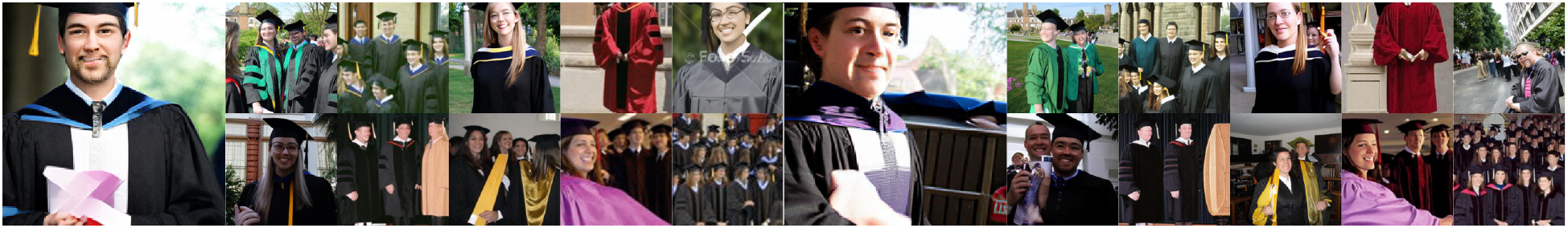}\\
{\scriptsize\itshape academic gown}\\[3pt]
\includegraphics[width=\textwidth]{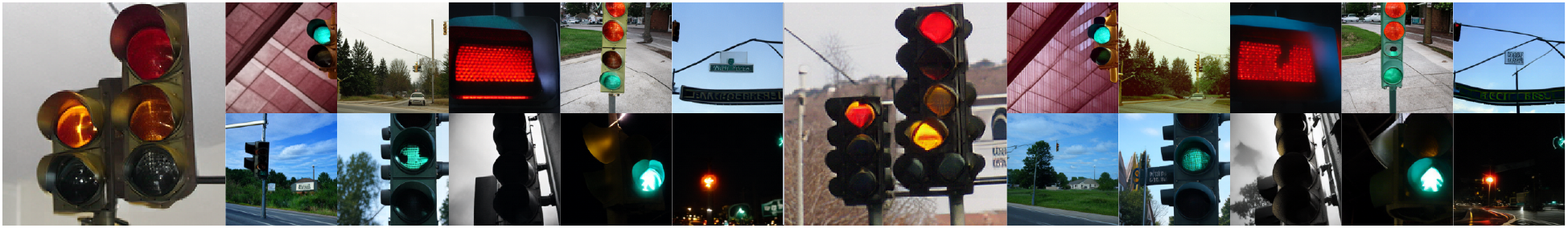}\\
{\scriptsize\itshape traffic light}\\[3pt]
\includegraphics[width=\textwidth]{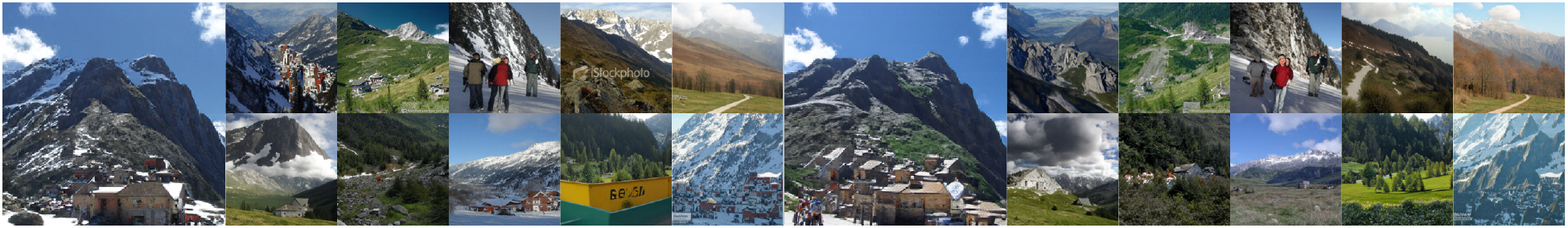}\\
{\scriptsize\itshape alp}
\caption{\textbf{Uncurated one-step samples} from \method{} and pMF-H FD-SIM \citep{fdloss} on five ImageNet-256 classes; column headers name the method. The two are close by eye despite the \swmetric{} gap of $1.30$ against $2.05$ in Table~\ref{tab:sw14}.}
\label{fig:qualitative}
\end{figure}

Figure~\ref{fig:qualitative} places uncurated \method{} samples beside pMF-H FD-SIM \citep{fdloss}, the strongest external one-step baseline by \swmetric{}, across five ImageNet-256 classes spanning a bird, an animal coat, a deformable garment, a rigid man-made object, and a natural landscape. Within each method the enlarged image is one sample and the adjacent $2\times5$ grid holds ten further draws under the same class label, taken as the first released draws without cherry-picking. Per sample the two models are hard to separate by eye, both reaching sharp, on-class images; this is precisely why a distributional metric is needed, as the \swmetric{} separation of $1.30$ against $2.05$ in Table~\ref{tab:sw14} is not visible in any single row.

\end{document}